\documentclass[runningheads]{llncs}
\usepackage[mobile]{eccv}

\usepackage{wrapfig}
\usepackage{graphicx}

\usepackage{algorithm}
\usepackage{algpseudocode}
\usepackage{enumerate}
\usepackage{enumitem}  %

\newcommand{\model}{RealWonder\xspace}

\newcommand{\myparagraph}[1]{\vspace{0.1cm}\noindent\textbf{#1}}
\renewcommand{\paragraph}[1]{\vspace{0.1cm}\noindent\textbf{#1}}

\definecolor{MyDarkBlue}{rgb}{0,0.08,1}
\definecolor{MyAqua}{rgb}{0,0.7,0.7}
\definecolor{MyDarkGreen}{rgb}{0.02,0.6,0.02}
\definecolor{MyDarkRed}{rgb}{0.8,0.02,0.02}
\definecolor{MyDarkOrange}{rgb}{0.40,0.2,0.02}
\definecolor{MyPurple}{RGB}{111,0,255}
\definecolor{MyRed}{rgb}{1.0,0.0,0.0}
\definecolor{MyGold}{rgb}{0.75,0.6,0.12}
\definecolor{MyDarkgray}{rgb}{0.66, 0.66, 0.66}

\definecolor{Cardinal}{rgb}{0.549,0.082,0.082}

\usepackage{eccvabbrv}
\usepackage{graphicx}
\usepackage{booktabs}
\usepackage[accsupp]{axessibility}
\usepackage[
    colorlinks=true,
    linkcolor=magenta,
    urlcolor=magenta,
    citecolor=eccvblue
]{hyperref}
\usepackage{orcidlink}
\usepackage{indentfirst}

\begin{document}

\title{RealWonder: Real-Time \\Physical Action-Conditioned Video Generation} 

\titlerunning{RealWonder}

\author{
\centering
Wei Liu$^{1*}$\quad
Ziyu Chen$^{1*}$\quad
Zizhang Li$^{1}$\\[0.2em]
Yue Wang$^{2}$\quad
Hong-Xing Yu$^{1\dagger}$\quad
Jiajun Wu$^{1\dagger}$\\[0.4em]
\small \url{https://liuwei283.github.io/RealWonder}\vspace{-0.5em}
}

\authorrunning{W. Liu, Z. Chen et al.}

\institute{
\centering
$^{1}$Stanford University\quad
$^{2}$University of Southern California\\[0.3em]
$^{*}$Equal contribution\quad
$^{\dagger}$Equal advising
}

\maketitle

\begin{center}
    \centering
    \captionsetup{type=figure}
    \includegraphics[trim={0px 0px 0px 0px}, clip, width=\linewidth]{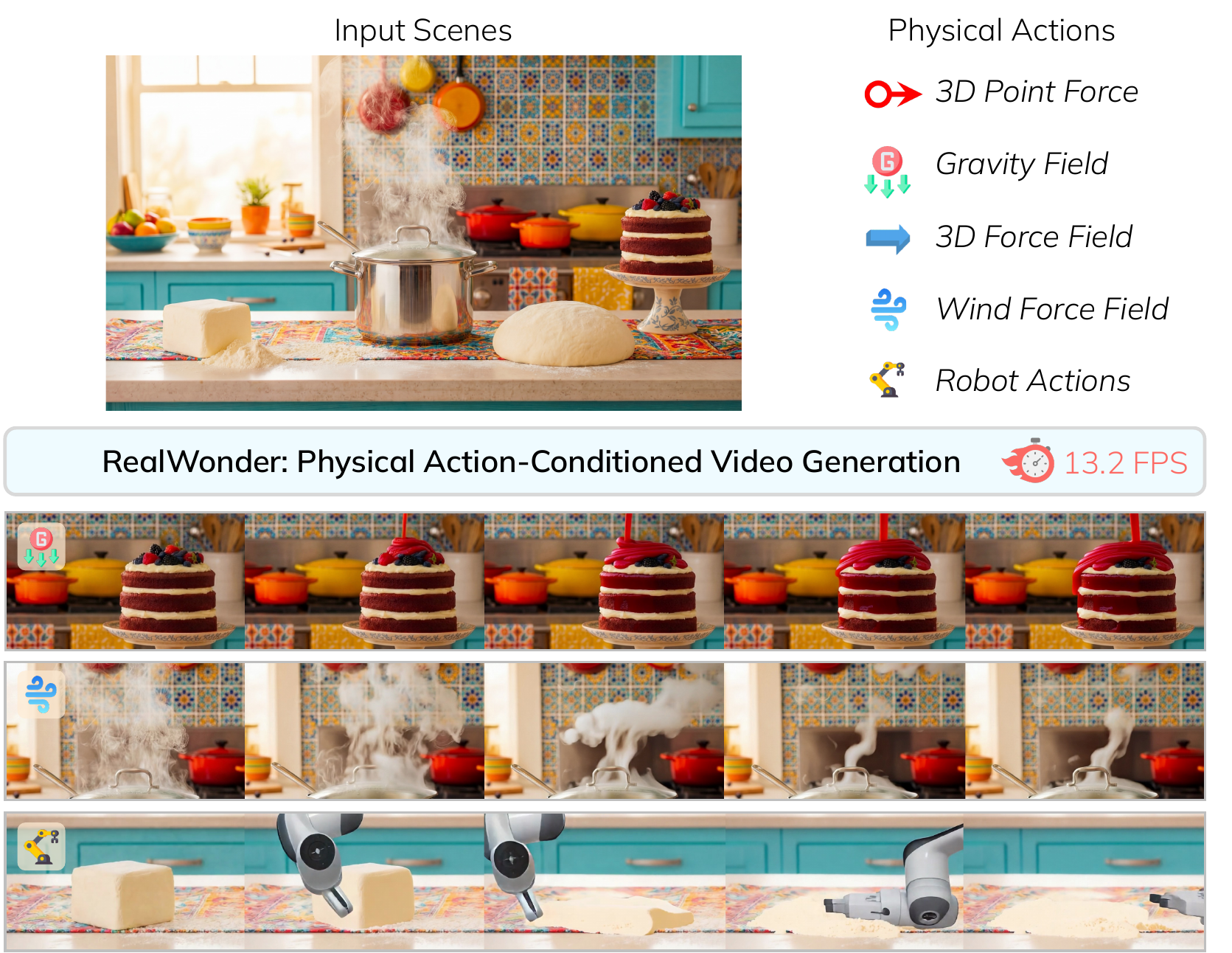}
    \caption{We propose \textbf{\model}, a real-time video generation model that is capable of simulating the consequences of 3D physical actions such as 3D forces, force fields, and robot gripper actions.
    }
    \label{fig:teaser}
\end{center}

\begin{abstract}
Current video generation models cannot simulate physical consequences of 3D actions like forces and robotic manipulations, as they lack structural understanding of how actions affect 3D scenes. We present RealWonder, the first real-time system for action-conditioned video generation from a single image. Our key insight is using physics simulation as an intermediate bridge: instead of directly encoding continuous actions, we translate them through physics simulation into visual representations (optical flow and RGB) that video models can process. RealWonder integrates three components: 3D reconstruction from single images, physics simulation, and a distilled video generator requiring only 4 diffusion steps. Our system achieves 13.2 FPS at 480$\times$832 resolution, enabling interactive exploration of forces, robot actions, and camera controls on rigid objects, deformable bodies, fluids, and granular materials. We envision \model opens new opportunities to apply video models in motion planning, AR/VR, and robot learning. Our code and checkpoints are publicly available on
\url{https://liuwei283.github.io/RealWonder}.

\keywords{3D Physical Actions \and Real-time Future Video Prediction}

\vspace{-0.5em}

\end{abstract}
    
\section{Introduction}
\label{s:intro}

The ability to interactively explore ``what-if'' scenarios with physical actions in real-time is fundamental to robotics simulation and AR/VR experiences. While recent advances in video generation have achieved impressive visual quality~\cite{wan2025,agarwal2025cosmos,2024sora} and even real-time performance~\cite{yin2025causvid,huang2025self,lin2025autoregressive} through acceleration techniques~\cite{yin2024improved,lin2025diffusion}, they remain fundamentally limited to passive generation or simple 2D controls~\cite{he2025matrix,ye2025yan}. The missing piece is the ability to condition these models on true 3D physical actions, e.g., forces, torques, and robot actions, and predict the future consequences caused by these actions.

The core challenge lies in the fundamental mismatch between how physical actions operate and how video models process information. Video diffusion models excel at understanding visual patterns in pixel or latent spaces, but lack the structural understanding to interpret how 3D forces propagate through 3D scenes. While recent video generation methods have explored conditioning through drag-based controls~\cite{yin2023dragnuwa,wu2024draganything} or motion trajectories~\cite{wang2024motionctrl,geng2024motion,shi2024motion}, these approaches remain computationally expensive and operate on 2D pixel space. Meanwhile, attempts~\cite{gillman2025forcepromptingvideogeneration,alhaija2025cosmostransfer,tu2025PlayerOne} to directly encode physical actions as tokens face two critical obstacles: first, actions like forces and torques are continuous and unbounded, resisting the tokenization schemes that work for discrete inputs like camera poses; second, obtaining action-video pairs for training remains an open problem, as inferring the precise physical actions that caused observed motions in videos is often infeasible.

We address these challenges by introducing physics simulation as an intermediate representation bridge between physical actions and video generation. We leverage physics simulators to translate action consequences into visual motion patterns that video models can naturally process. Specifically, given 3D physical actions, we first predict their consequences through physics simulation, rendering the results as optical flow and coarse RGB previews. These intermediate representations preserve the causal relationship between actions and outcomes while providing the visual signals that modern video generators are designed to process. This approach elegantly sidesteps the tokenization problem by turning continuous action signals into discrete pixels with physics simulators. What makes this method scalable is that it does not require action-video pairs, and only 2D flow-video pairs are needed to train the video generator. 

Building on such intuition, we introduce \textbf{RealWonder}, a physics-aware system enabling \textbf{real-time physical action-conditioned video generation} from a single image. RealWonder integrates three carefully designed components: a 3D scene reconstruction module that lifts 2D input images into simulatable 3D representations, a physics simulator that predicts action outcomes, and a distilled flow-conditioned four-step video generator that synthesizes photorealistic results in real-time. Given an input image and a sequence of physical actions (e.g., forces applied to objects, robot actions, and camera motions), RealWonder first reconstructs the underlying 3D scene geometry and estimates material properties. The physics simulator then computes the dynamic evolution under the specified actions, producing flow fields and coarse RGB renderings. These physics-grounded representations, combined with the original image, condition our distilled video model to produce realistic video streaming at interactive speeds, achieving up to 13.2 FPS at 480$\times$832 resolution on a single GPU.

Our experiments demonstrate that RealWonder successfully bridges the gap between physical understanding and visual synthesis, enabling real-time interaction with diverse materials including rigid objects, deformable bodies, fluids, and granular substances. Users can apply forces, control robotic grippers (see Figure~\ref{fig:teaser} for example), or move cameras, and immediately see realistic visual consequences unfold. In summary, our contributions are threefold:
\begin{itemize}
    \item We present \textbf{RealWonder}, the first real-time streaming system for action-conditioned video generation that accepts \textbf{3D physical actions} including forces, torques, robot actions, and camera control as input, achieving up to 13.2 FPS generation at 480$\times$832 resolution.
    \item We introduce a novel architecture that uses physics simulation as an intermediate representation, enabling the prediction of 3D physical action consequences without training on action-video pairs or tokenizing actions.
    \item We design a distillation scheme that incorporates optical flow conditioning into video generation, significantly reducing diffusion overhead while achieving effective flow control. 
\end{itemize}

\section{Related Work}%

\myparagraph{Controllable Video Generation}.
Recent advances in video diffusion models have motivated extensive research into adding user control, spanning through 2D control like depth~\cite{agarwal2025cosmos,alhaija2025cosmostransfer}, sketches~\cite{xing2024tooncrafter}, drag-based interaction~\cite{yin2023dragnuwa,wu2024draganything} for screen-space trajectories, subject-specific control~\cite{hu2023animateanyone}, optical flow-based methods~\cite{niu2024mofa,burgert2025go} for dense pixel-level specification, etc.
Spatial control has emerged as particularly promising, evolving through camera control~\cite{he2024cameractrl,zheng2024cami2v,li2025hunyuangamecrafthighdynamicinteractivegame,he2025matrix}, trajectory-based approaches~\cite{wang2024motionctrl,zhang2024tora} for complete motion paths. Despite impressive results~\cite{geng2024motion,shi2024motion,gillman2025forcepromptingvideogeneration}, these approaches operate in 2D screen space and require complete motion specifications before generation begins, preventing their application to scenarios requiring 3D physical understanding such as predicting force propagation or simulating robotic manipulations.

\myparagraph{Real-time and Streaming Video Generation}.
The computational demands of video diffusion models have driven significant advances in acceleration techniques. Recent works employ adversarial post-training~\cite{lin2025autoregressive} or Distribution Matching Distillation~\cite{yin2025causvid} to train causal generators with one- or few-step diffusion for streaming generation. Follow-up works~\cite{huang2025self,liu2025rolling,yang2025longlive} mitigate the drifting issue inherent in causal generators to improve temporal consistency. A concurrent work, MotionStream~\cite{motionstream2025}, also achieves real-time streaming generation with trajectory-based control. However, these approaches cannot accept 3D physical actions as input to \emph{predict} the future evolution, as their focus on video generation with user-provided control signals such as text and trajectory.

\myparagraph{Interactive Video World Models}.
The vision of interactive world models~\cite{ha2018world} extends beyond passive generation to enable active user participation, with recent work demonstrating impressive results in closed domains like video game environments~\cite{bruce2024genie,che2024gamegen,valevski2024diffusion} where action-video pairs are readily available. Recently, finetuning base video models with pairwise action-video data from synthetic engines also shows promising generalization to real image inputs, such as LingBot-World~\cite{team2026advancing} and others~\cite{he2025matrix,mao2025yume,sun2025worldplay}. However, extending the action space to physical actions faces fundamental obstacles~\cite{bar2024navigation,agarwal2025cosmos}, unlike synthetic environments with discrete action sets, 3D physical actions are continuous, high-dimensional, and unbounded. For example, forces vary in magnitude, direction, and point of application, and robotic actions often involve complex kinematic chains. Current world models lack architectural components to process such actions directly, limiting interactive video generation to simplified actions such as camera movement.

\myparagraph{Physics-based Video Generation}.
Physics-based methods reconstruct 3D scenes, apply physics solvers, and render results, expanding from early work on object vibrations~\cite{davis2015visual} to recent systems handling diverse materials including rigid objects~\cite{le2023differentiable,chen2025physgen3d}, deformable objects~\cite{chen2022virtual,li2023pac,xie2024physgaussian,zhang2024physdreamer,jiang2025phystwin,liu2024physics3d,huang2024dreamphysics} and fluids~\cite{gao2025fluidnexus}.
WonderPlay~\cite{li2025wonderplay} represents prior work that combines physics simulation with video generation, though requiring several minutes for short clips. We draw inspiration from it but with an important distinction: Instead of slow optimization of explicit 4D representations to render videos, we leverage physics simulation as an intermediate bridge to interface a video generator with a novel distillation scheme, and thus we can enable real-time physical action-conditioned video generation with high visual realism.

\vspace*{-4pt}\section{\model}

\begin{figure*}[t]
  \centering
   \includegraphics[width=\linewidth]{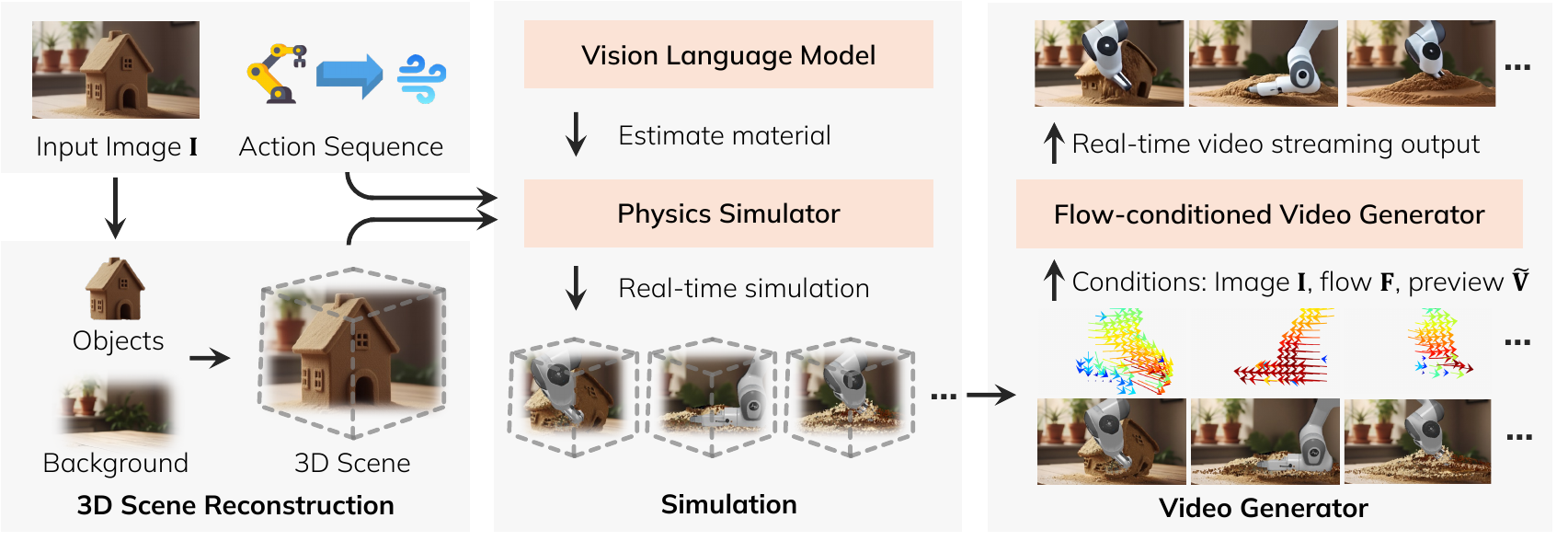}

   \caption{\textbf{Overview of \model.} (\textbf{Left}) Given a single image and a sequence of actions as input, we first reconstruct the 3D scene as point clouds, (\textbf{Middle}) estimate material for the objects to interact with, and then maintain a physics simulation stream using the actions. (\textbf{Right}) Meanwhile, we maintain another stream of rendering optical flow and RGB preview to condition a few-step conditional video generator, producing the physical action-conditioned video streaming.}
   \label{fig:approach}
\end{figure*}

\myparagraph{Formulation.} Our goal is real-time generation of physically plausible videos from 3D actions. The input consists of a single RGB image $\mathbf{I} \in \mathbb{R}^{H \times W \times 3}$ and a sequence of 3D physical actions $\{\mathbf{a}_t\}, t=1,2,\cdots$. These actions encompass three categories: (1) external forces $\mathbf{f}_t(x,y,z) \in \mathbb{R}^3$ applied at 3D positions, (2) robot end-effector commands $\mathbf{r}_t = \{\mathbf{p}_t^\text{ee}, \mathbf{q}^\text{ee}_t, g_t\}$ specifying position, orientation, and gripper state for a pre-defined robot model (e.g., Franka), and (3) camera poses $\mathbf{C}_t = \{\mathbf{R}_t, \mathbf{t}_t\}$ defining viewpoint rotation and translation. The output is a video stream $\mathbf{V}_t$ generated at interactive rates (up to 13.2 FPS at 480$\times$832 resolution) that depicts realistic visual consequences of applying these actions to the scene. The fundamental challenge lies in bridging the semantic gap between continuous 3D physical actions (which operate in force and kinematic spaces) and video generation models that process visual patterns in pixel or latent spaces.

\myparagraph{Overview.} RealWonder addresses this challenge through a three-stage pipeline that uses physics simulation as an intermediate representation bridge, illustrated in Figure~\ref{fig:approach}. First, we reconstruct a simulatable 3D scene representation from the input image, estimating geometry and material properties suitable for real-time physics (Section~\ref{method:3drecon}). Second, we apply physics simulation to compute the scene's dynamic response to input actions, rendering the results as optical flow $\mathbf{F}_t \in \mathbb{R}^{H \times W \times 2}$ and coarse RGB previews $\tilde{\mathbf{V}}_t \in \mathbb{R}^{H \times W \times 3}$ that encode motion patterns while preserving action causality (Section~\ref{method:physics_sim}). These physics-grounded visual representations, combined with the original image, condition a distilled video generator that produces photorealistic results in 4 diffusion steps (Section~\ref{method:vidgen}). This architecture elegantly sidesteps two fundamental obstacles: it avoids tokenizing continuous actions by leveraging physics simulators that naturally handle unbounded force and action inputs, and it eliminates the need for scarce action-video training pairs by training only on flow-video correspondences. The complete system achieves sub-100ms latency per frame through careful co-design of all components for streaming generation (Section~\ref{sec:streaminfer}).

\subsection{Single-Image 3D Scene Reconstruction}\label{method:3drecon}

Our 3D scene representation $\mathcal{S} = \mathcal{B} \cup \mathcal{O}$ consists of a static background $\mathcal{B}$ and dynamic objects $\mathcal{O}$. For real-time performance, we employ lightweight point cloud representations that provide sufficient detail of geometry while minimizing computational overhead during rendering.

\myparagraph{Background.} We represent the background as a point cloud $\mathcal{B} = \{(\mathbf{p}_i^B, \mathbf{c}_i^B)\}_{i=1}^{N_B}$ where $\mathbf{p}_i^B \in \mathbb{R}^3$ denotes 3D positions and $\mathbf{c}_i^B \in \mathbb{R}^3$ denotes RGB colors. The background is reconstructed by segmenting static regions, inpainting areas occluded by objects, estimating per-pixel depth, and unprojecting to 3D space. These points serve as static collision boundaries during simulation.

\myparagraph{Objects.} An ``object'' refers to any dynamic entity we simulate in the physics solver, including rigid bodies, cloth, granular materials, and fluids. We represent objects as point clouds: $\mathcal{O} = \{(\mathbf{p}_j^O, \mathbf{c}_j^O, \mathbf{v}_j)\}_{j=1}^{N_O}$, where $\mathbf{v}_j \in \mathbb{R}^3$ denotes velocity. For each segmented object, the points are derived from unprojected pixels similar to the background and are supplemented with mesh vertices for invisible surfaces. To do this, we generate a complete 3D mesh for an object using a feed-forward reconstruction model. The meshes are then registered to the scene coordinate frame through pose estimation and scale alignment. We extract mesh vertices of invisible surfaces (e.g., the back of the object) to complement the unprojected object pixels, ensuring we capture the complete object geometry necessary for accurate physics simulation and rendering. 

\myparagraph{Materials.} We employ a vision-language model (VLM) to classify each object into one of six material categories (rigid, elastic, cloth, smoke, liquid, granular) and estimate corresponding physical parameters $\mathbf{m}$~\cite{li2025wonderplay}. These include density, friction coefficients, elastic moduli, and viscosity, depending on the material type. While the VLM provides reasonable initial estimates, users can override these classifications and parameters for creative control or improved physical accuracy.

We leave implementation details of reconstruction and material estimation in the Appendix.

\subsection{Physics Simulation as Intermediate Bridge}\label{method:physics_sim}

Physics simulation provides the critical bridge between 3D physical actions and video generation by translating physical 3D actions into visual motion patterns (optical flow and RGB frames) that video models can naturally process.

\myparagraph{Action Representation.} We unify three types of actions in 3D scene space. External forces $\mathbf{f}_t(x,y,z)$ are directly applied at specified 3D positions. Robot end-effector commands $\mathbf{r}_t = \{\mathbf{p}_t^\text{ee}, \mathbf{q}^\text{ee}_t, g_t\}$ specifying position, orientation, and gripper state are converted to joint torques and forces through solving inverse kinematics (IK), which then drive the articulated robot model in simulation~\cite{Genesis}. Camera poses $\mathbf{C}_t = \{\mathbf{R}_t, \mathbf{t}_t\}$ are applied during rendering.

\myparagraph{Physics Solvers.} At each timestep $t$, our physics engine takes the current scene state $\mathcal{S}_t$ and actions $\mathbf{a}_t$ as input, computing updated positions $\mathbf{p}_{t+1}$ and velocities $\mathbf{v}_{t+1}$ for all dynamic points:
\begin{equation}
(\mathbf{p}_{t+1}, \mathbf{v}_{t+1}) = \texttt{PhysicsStep}(\mathcal{S}_t, \mathbf{a}_t),
\end{equation}
where we employ specialized solvers for different materials: rigid body dynamics through shape matching for collision handling~\cite{muller2005meshless}, Position-Based Dynamics (PBD) for elastic bodies, cloth and smoke~\cite{bender2015position}, Material Point Method (MPM) for liquid and granular materials~\cite{jiang2016material}. These solvers are coupled to handle multi-material interactions. A single physics step typically completes in less than 2ms.

\myparagraph{Intermediate Representations.} The physics simulation output is transformed into two visual representations that serve as conditioning signals for video generation:

\vspace{0.1cm}
\noindent
\underline{Optical Flow.} We compute pixel-space flow $\mathbf{F}_t \in \mathbb{R}^{H \times W \times 2}$ by projecting the 3D velocity field:
\begin{equation}
\mathbf{F}_t(u,v) = \Pi(\mathbf{p}_t + \Delta t \cdot \mathbf{v}_t) - \Pi(\mathbf{p}_t),
\end{equation}
where $\Pi$ denotes camera projection and $(u,v)$ are pixel coordinates. This flow field captures the motion consequences of applied actions.

\vspace{0.1cm}
\noindent
\underline{Coarse RGB Rendering.} We render a preview video $\tilde{\mathbf{V}}_t \in \mathbb{R}^{H \times W \times 3}$ using simple point cloud rasterization. While visually approximate, this preview provides crucial structural cues such as occlusion changes that pure flow cannot capture.

These intermediate representations achieve three critical goals: (1) they preserve the causal relationship between actions and their visual consequences, (2) they exist in the visual domain where video models can process, and (3) they can be computed in real-time.

\vspace{-0.2cm}

\subsection{Real-Time Conditional Video Generation}\label{method:vidgen}
\begin{figure*}[t]
  \centering
  \includegraphics[width=\linewidth]{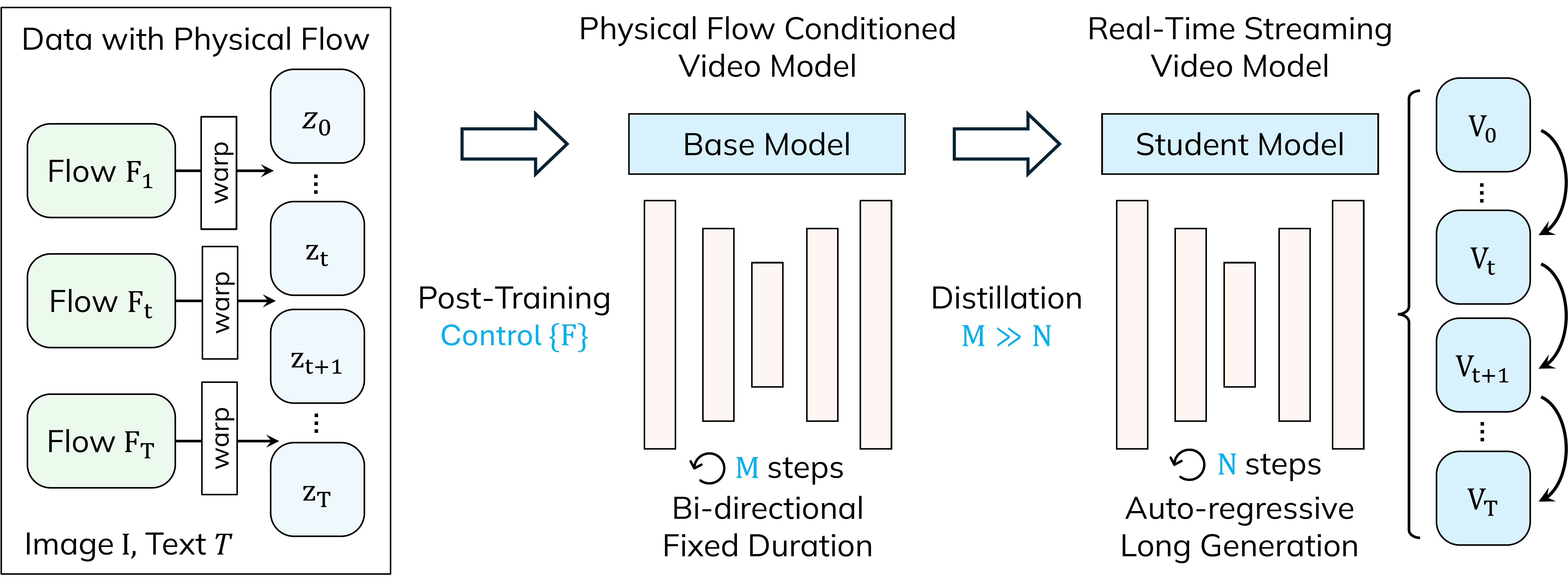}
  \caption{\textbf{Training of Real-time Flow-Conditioned Video Model.} A pretrained image-to-video model~\cite{wan2025} is first adapted to optical-flow conditioning through LoRA~\cite{hu2022lora} post-training. Next, it is distilled via distribution-matching training into a causal, flow-conditioned real-time video generator.}
  \label{fig:videotrain}
\end{figure*}

Given the physics-derived optical flow $\mathbf{F}_t$ and coarse preview $\tilde{\mathbf{V}}_t$, we require a video generator that can turn these cues into photorealistic results at interactive speeds. Our module $\mathcal{G}$ takes the input image $\mathbf{I}$, text prompt $\texttt{text}$, and these conditions to produce physically plausible video in real time: $\mathbf{V}_t=\mathcal{G}(\texttt{text}, \mathbf{I}, \mathbf{F}_t, \tilde{\mathbf{V}}_t)$. Here we describe how the model is trained to take flow conditioning, the RGB preview $\tilde{\mathbf{V}}_t$ will be used at inference time detailed in Section~\ref{sec:streaminfer}. While modern video diffusion models excel at generating high-fidelity videos, they require many denoising steps (typically 50) and process a sequence of frames in parallel, preventing real-time interaction. We address these challenges through a two-stage approach: (1) augmenting a pretrained image-to-video model with flow conditioning, then (2) distilling this multi-step teacher into a 4-step causal student capable of streaming generation. The pipeline is illustrated in Figure~\ref{fig:videotrain}.

\myparagraph{Flow-Conditioned Teacher Model.} We start from a pretrained image-to-video diffusion model $\mathcal{G}_\text{base}(\texttt{text},\mathbf{I})$~\cite{wan2025,videoxfun2024} and augment it with flow control through post-training. Given a training video $\mathbf{V}_t$ with extracted optical flow $\mathbf{F}_t$, we employ flow-based noise warping~\cite{burgert2025go}: we sample single-frame Gaussian noise $z\sim\mathcal{N}(0,I)$ and warp it temporally according to the flow field to obtain structured noise $z^\mathbf{F}=\text{Warp}(z, \mathbf{F})$. This warping preserves the Gaussian distribution while encoding motion patterns directly into the noise structure~\cite{burgert2025go}. We then finetune $\mathcal{G}_\text{base}$ using the flow-matching objective to model the velocity field between flow-warped noise and the data distribution. Unlike prior motion-controlled methods~\cite{geng2024motion,agarwal2025cosmos} that require additional embedding modules or architectural changes, this approach injects control directly through the initial noise, maintaining efficiency while achieving precise motion adherence.

\myparagraph{Causal Distillation for Streaming.} The flow-conditioned teacher remains a bidirectional model requiring full sequence processing, incompatible with real-time streaming generation. We distill it into a causal student that generates frames sequentially in just 4 denoising steps. Following Distribution Matching Distillation (DMD)~\cite{yin2024one,yin2024improved}, we minimize the reverse KL divergence between the student's output distribution and the teacher's:
\begin{equation}
\nabla L_{\text{DMD}}
= \mathbb{E}_{t}\!\left[
\nabla_{\theta}\, \mathrm{KL}\!\left(
p_{\text{fake}, t} \,\|\, p_{\text{real}, t}
\right)
\right].
\end{equation}
where $p_{\text{fake}, t}$ denotes the student’s estimated distribution at time $t$. 
To enable stable long-horizon generation, we adopt the Self Forcing training paradigm~\cite{huang2025self} with autoregressive rollout during training. However, standard Self Forcing exhibits quality degradation for extended sequences. We fix this by storing the KV cache before RoPE~\cite{su2021roformer} is applied and adding attention sink, similar to concurrent works~\cite{motionstream2025,liu2025rolling,infinite-forcing}.

\vspace{-0.2cm}

\subsection{Streaming Inference}\label{sec:streaminfer}

\vspace{-0.2cm}

\myparagraph{RGB Conditioning via SDEdit.} While our distilled model is trained with flow conditioning, it is straightforward to incorporate the coarse RGB preview $\tilde{\mathbf{V}}_t$. We incorporate this additional signal through SDEdit~\cite{meng2021sdedit} during the 4-step denoising process. Specifically, instead of denoising from pure flow-warped noise $\mathbf{V}_{t,(4)} = z^{\mathbf{F}}_t$ all the way to the clean sample $\mathbf{x}_0$, we begin denoising from a mixture:
\begin{equation}
\mathbf{V}_{t,(3)} = \alpha_{(3)} \cdot \mathcal{E}(\tilde{\mathbf{V}}_t) + \sqrt{1-\alpha_{(3)}^2} \cdot z_t^\mathbf{F},
\end{equation}
where $\mathcal{E}$ is the VAE encoder, $\alpha_{(3)}$ is the noise schedule coefficient at diffusion step 3. By initiating denoising from step 3 rather than step 4, we allow the model one initial step to denoise the mixture before performing the remaining three standard denoising steps. This dual conditioning preserves the motion accuracy from flow while incorporating structural cues from the physics preview, resulting in coherent object deformations and occlusion handling.

\myparagraph{Streaming Architecture.} Our system maintains two parallel streams: (1) physics simulation and its rendering into intermediate representations, and (2) video generation operating at 13.2 FPS, consuming the latest physics-derived conditioning. The causal nature of our distilled model $\mathcal{G}$ enables frame-by-frame generation, expressed as:
\begin{equation} 
\mathbf{V}_{t+1}
= \mathcal{G}(\texttt{text},\mathbf{I},\mathbf{F}_{t+1},\tilde{\mathbf{V}}_{t+1},\{\mathbf{V}_j\}_{j\leq t}).
\end{equation}
\myparagraph{Pipeline Summary.} At inference time, given an input image $\mathbf{I}$ and a stream of user actions $\mathbf{a}_t$, we first reconstruct the 3D scene, then continuously run physics simulation to generate optical flow $\mathbf{F}_t$ and coarse RGB previews $\tilde{\mathbf{V}}_t$. These physics-derived representation streams condition our distilled video generator to produce a photorealistic frame stream. We summarize it in Alg.~\ref{alg:pipeline} in Appendix.

\vspace{-0.2cm}

\section{Experiments}%
\label{s:experiments}

\myparagraph{Implementation details.} 
For physics simulation, we adopt the Genesis as the simulator~\cite{Genesis}. Our simulation uses a time step of 0.01s with up to 20 substeps per simulation step for numerical stability. The simulation and rendering stream operates at 30 FPS in parallel to the video generator stream. For robot actions, we use the Franka robot model provided by Genesis and support interactions between robots and multiple materials.
For video model training, we adopt the VideoXFun~\cite{videoxfun2024} wan2.1-1.3B-InP model as the I2V base model. We freeze all its weights and inject LoRA~\cite{hu2022lora} modules into every attention block with the rank of 2048. The model is then trained for 300K iterations with learning rate $10^{-5}$ to learn flow conditioning. After this stage, we apply Self Forcing~\cite{huang2025self} style training to obtain a distilled, real-time, flow-conditioned video generator. For the causal adaptation stage of distillation~\cite{yin2025causvid}, we sample 2K ODE trajectories from the post-trained model mentioned above and train the student model for 3K iterations with the MSE loss. We then apply distribution matching distillation for 600 iterations. The total training compute was approximately 128 A100 GPU-days.
We leave further implementation details in the Appendix.

\myparagraph{Dataset.} To train our video generator, we follow~\cite{burgert2025go} and construct 200K pairs of flow-warped-noise and video using RAFT~\cite{teed2020raft}. Among them, 180K videos are real-world clips from OpenVid~\cite{nan2024openvid}, filtered to 80-120 frames, and the remaining 20K are synthetic videos generated by the Wan2.1-14B-T2V~\cite{wan2025} model using prompts from VidProM~\cite{wang2024vidprom}.

\myparagraph{Baselines.} We compare against three open-source representative baselines: For physics-based video generation, we compare to PhysGaussian~\cite{xie2024physgaussian} which models 3D scene dynamics by integrating MPM~\cite{jiang2016material} simulation with reconstructed 3D Gaussian Splatting particles. We provide our reconstructed 3D points as initialization and run optimization with the input image. 
For conditional video generation methods, we compare against CogVideoX-I2V~\cite{yang2024cogvideox} which is a state-of-the-art open-source video model~\cite{duan2025worldscore} with text and image as conditions, and Tora~\cite{zhang2024tora} which additionally allows drag-based conditioning. For Tora, we use the trajectories from our simulation as the drag input.

\begin{figure*}
    \centering
    \includegraphics[width=\linewidth]{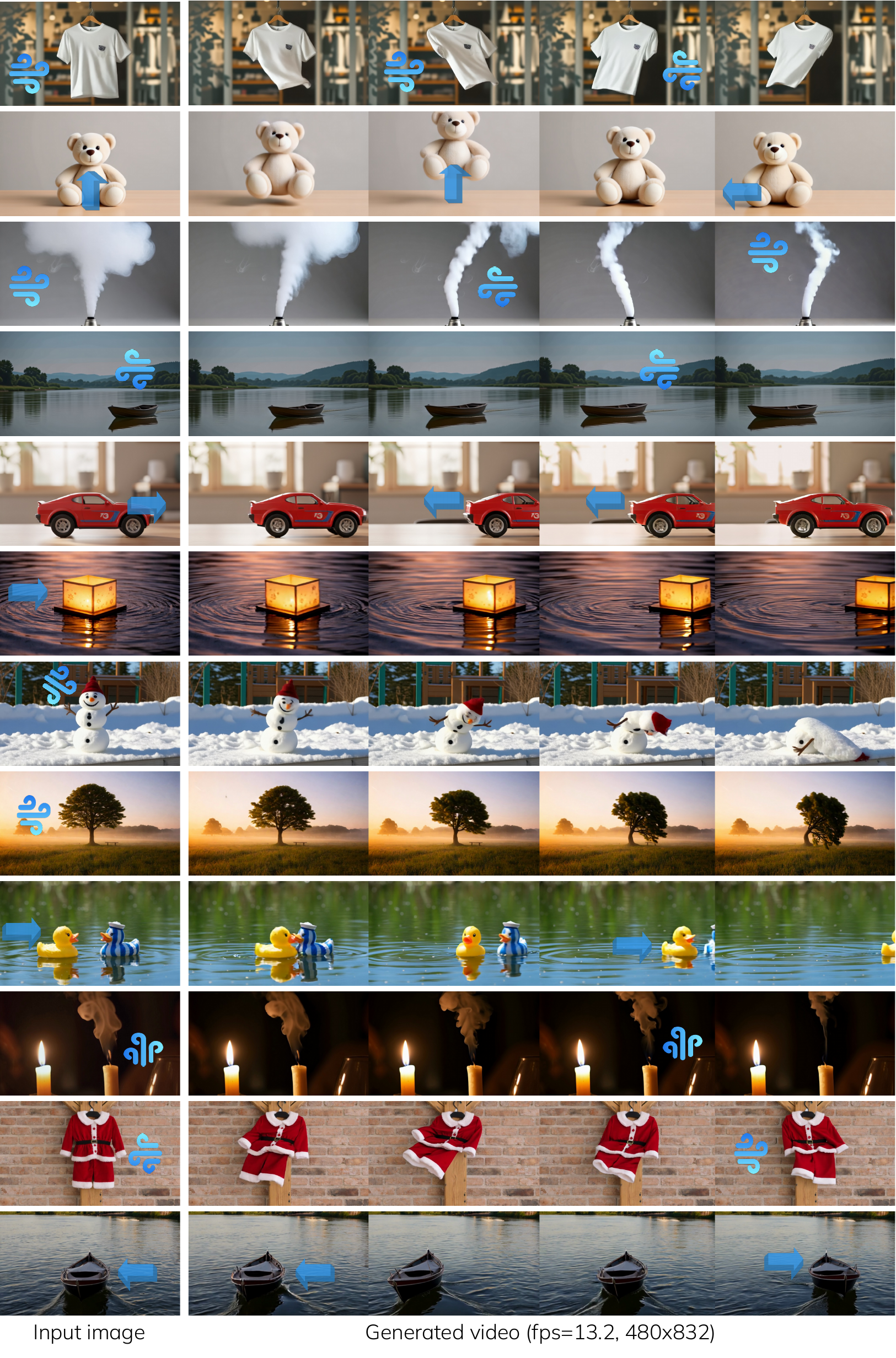}
    \caption{\textbf{Qualitative results}. In the left column we show the input scene image and initial actions, where the arrow indicates 3D force and the blue wind icon indicates force fields. Note that we always apply gravity in our simulation.}
    \label{fig:qualitative}
\end{figure*}

\begin{figure}[t]
    \centering
    \includegraphics[width=\linewidth]{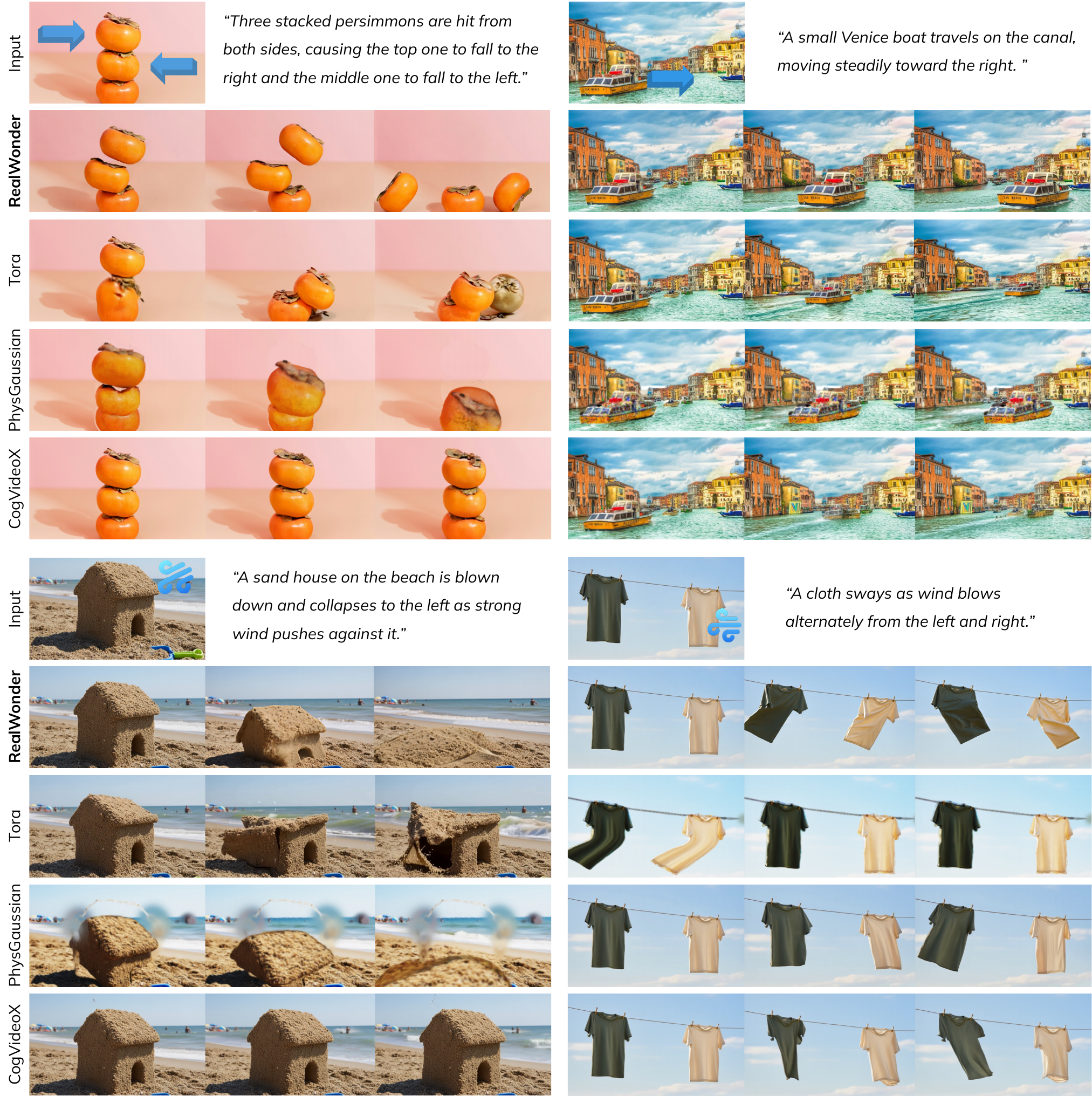}
    \vspace{-0.5em}
    \caption{\textbf{Comparison to baselines}. In the first row, we show the input images and actions (we use arrows for 3D point forces and wind icon for force fields) together with the text prompts. We always apply gravity in our simulation.}
    \label{fig:comparison}
    \vspace{-2em}
\end{figure}

\myparagraph{Evaluation metrics.} 
Since physical 3D action-conditioned video generation is a new task, there is no standard dataset or benchmark for evaluation. Thus, we curate a new dataset of 30 images including real and synthetic images of diverse materials (e.g., cloth, rigid body, elastic objects, liquid, gas, sand, snow) and proper physical actions associated with each image. For quantitative metrics,
we adopt the imaging, aesthetic, and consistency metrics from VBench~\cite{huang2024vbench}. We also adopt the GPT-4o-based physical realism metric~\cite{chen2025physgen3d}. To measure human preference, we recruit 400 participants and conduct a user study. We employ a Two-alternative Forced Choice (2AFC) protocol. Each participant evaluates 6 test scenes for all 3 baseline comparisons. The participants view an action description alongside a randomly ordered side-by-side comparison video: one from our method and one from a baseline. Participants then select which video demonstrates superior performance in one of three criteria: \textit{Action following} which judges if the model follows the given physical action and predicts the consequence, \textit{physical plausibility} which measures the plausibility of the predicted motion, \textit{motion fidelity} that reflects the naturalness of the generated motion, and the subjective \textit{visual quality} of the videos.

\subsection{Results}
\begin{table}[t]
\centering
\setlength{\tabcolsep}{4pt}

\renewcommand{\arraystretch}{1.2}
\caption{Quantitative comparison to baselines.}
\vspace{-1em}
\begin{tabular}{lccccc}
\toprule
Methods & Visuals ($\uparrow$) & Aesthetics ($\uparrow$) & Consistency ($\uparrow$) & PhysReal ($\uparrow$) \\
\midrule
PhysGaussian & 0.454 & 0.517 & 0.221 & 0.468 \\
CogVideoX & 0.696 & \textbf{0.603} & \underline{0.234} & \underline{0.624} \\
Tora & \underline{0.700} & 0.588 & 0.223 & 0.578 \\
\model (ours) & \textbf{0.708} & \underline{0.593} & \textbf{0.265} & \textbf{0.705} \\
\bottomrule
\end{tabular}
\label{tab:quantitative}
\end{table}

\begin{table}[t]
\setlength{\tabcolsep}{2pt}
\renewcommand{\arraystretch}{1.2}
\centering
\caption{2AFC human study results of the favor rate of our \model over baseline methods. The responses are collected from 400 participants.}
\vspace{-1em}
\resizebox{\columnwidth}{!}{%
\begin{tabular}{lcccc}
\toprule
                        & \textbf{Action Following} & \textbf{Motion Fidelity} & \textbf{Visual Quality} & \textbf{Physical Plausibility} \\
\midrule
over PhysGaussian~\cite{xie2024physgaussian}  & 88.4\%  & 82.0\%  & 88.6\% & 87.1\%  \\
over CogVideoX-I2V~\cite{yang2024cogvideox}   & 89.6\%  & 71.0\%  & 75.3\% & 85.9\% \\
over Tora~\cite{zhang2024tora}                & 83.9\%  & 67.9\%  & 75.4\% & 79.7\% \\
\bottomrule
\end{tabular}}
\label{tab:user-study}
\vspace{-2em}
\end{table}

\myparagraph{Comparison to baselines.}
In Figure~\ref{fig:comparison}, We show qualitative visual comparisons with baseline methods on different examples. The top row shows input images, actions and the text prompt, followed by the video frames generated from \model and the baselines.

Despite their ability to produce plausible visual quality, baseline video models struggle to generate outcomes of the physical actions. In the persimmon example (top-left in Figure~\ref{fig:comparison}), Tora~\cite{zhang2024tora} fails to apply the hit on the middle persimmon. Furthermore, it violates object permanence, merging two persimmons and creates a new persimmon in the end. CogVideoX-I2V~\cite{yang2024cogvideox} fails to apply either hit and it struggles to generate any dynamics. Both models also struggle with the boat example (top-right in Figure~\ref{fig:comparison}). Although Tora is conditioned on our simulated 2D trajectory, it wrongly interprets it as moving forward rather than moving to the right. CogVideoX-I2V shows similar issue, perhaps because these video models are biased to drive a boat moving forward rather than predicting the future evolution according to the specified physical actions.

The physics-based method, PhysGaussian~\cite{liu2024physgen}, is able to synthesize reasonable boat motion that moves to the right. However, the visual quality is compromised as it relies completely on the 3D representations to create appearances, and it cannot synthesize dynamic shading and shadow.

Our approach, in contrast, offers the best of both worlds to synthesize realistic action-conditioned videos: the physics simulation provides the adherence to physical actions and the consistency to physical laws, and the video model provides visual realism such as dynamic shadows, water splashes, collapsing sand, and soft clothing.

We report quantitative results in Table~\ref{tab:quantitative} and user study results in Table~\ref{tab:user-study} using the 2AFC setup described above. \model achieves the best or second best numbers across all metrics, and it is significantly more favored compared to other baselines.

\myparagraph{Generation speed.} We report system speeds in Table~\ref{tab:runtime_comparison}. The baseline video models are not streamable but limited to a single time window of around 5 seconds, so we only compute FPS by diving the frame counts by the generation time. For \model and PhysGaussian, we compute both latency (the time between specifying an action and seeing the outcome) and FPS (frames per second during streaming). We evaluate all methods on a single H200 GPU, except PhysGaussian which runs on an A6000 GPU due to environment incompatibility. \model achieves real-time streaming generation.

\begin{table}[t]
    \centering
    \caption{Comparison of runtime performance.}
    \vspace{-0.5em}
    \setlength{\tabcolsep}{8pt}
    \begin{tabular}{lcccc}
        \toprule
         & Tora & CogVideoX-I2V & PhysGaussian & RealWonder (Ours) \\
        \midrule
        FPS     & 0.107 & 0.225 & 0.207 & \textbf{13.2} \\
        Latency & - & - & 4.84s & \textbf{0.73s} \\
        \bottomrule
    \end{tabular}
    \label{tab:runtime_comparison}
    \vspace{-1em}
\end{table}

\begin{figure}[t]
    \centering
    \includegraphics[width=\linewidth]{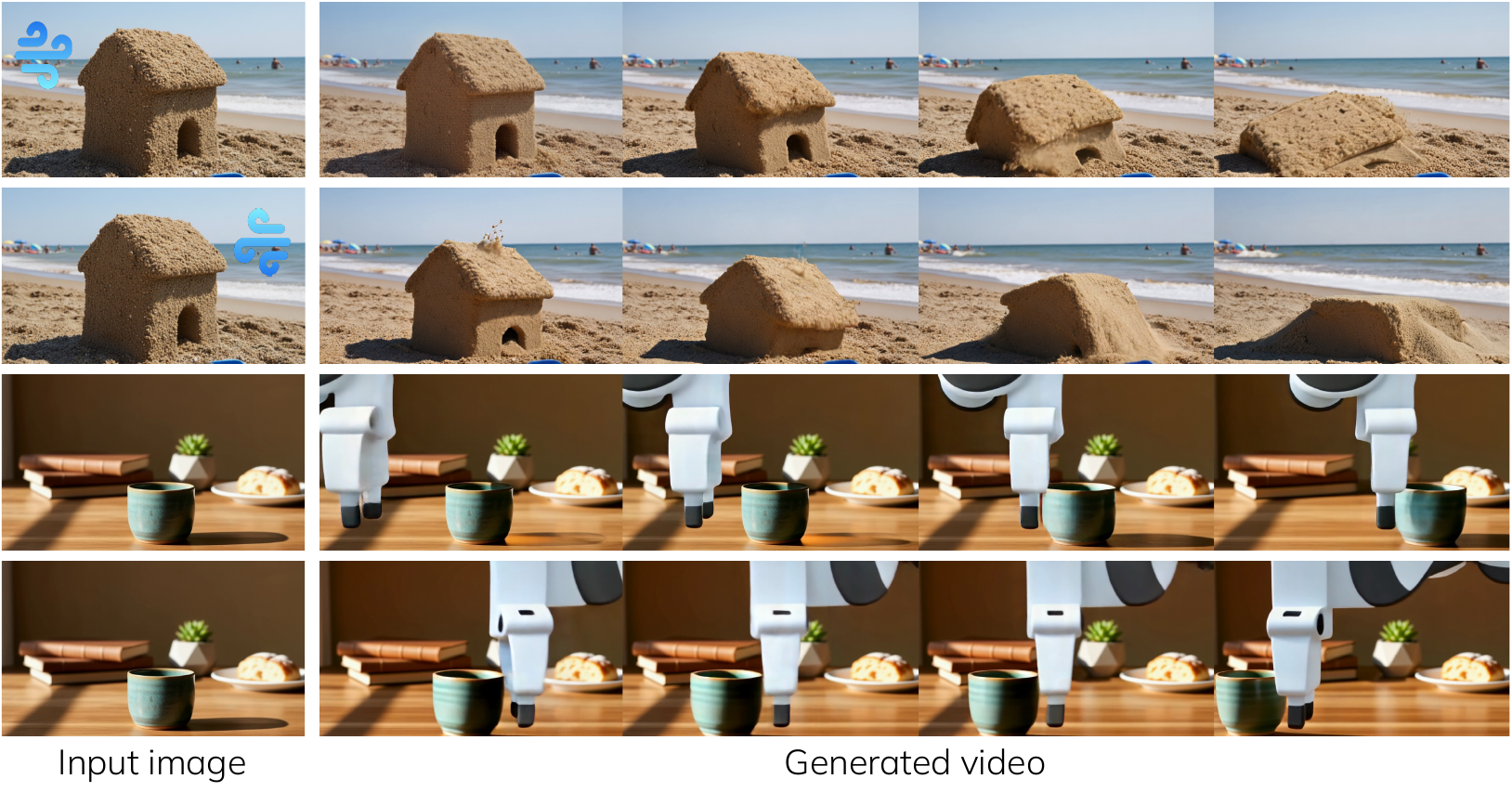}
    \caption{\textbf{Different actions} on the same scene, leading to different generated physical outcomes.}
    \label{fig:diff-interact}
    \vspace{-2em}
\end{figure}

\myparagraph{Different types of object material and actions.}
In Figure~\ref{fig:qualitative} and Figure~\ref{fig:teaser}, we show results on a variety of input images with diverse physical actions, including robot gripper actions, 3D point force, and 3D force fields (e.g., wind). These scenes include different types of materials, including rigid objects, deformable objects, cloth, fluids, and the interaction among them. These demonstrate \model's capability in simulating different actions on different materials.

\myparagraph{Different actions on the same scene.}
In Figure~\ref{fig:diff-interact}, we showcase that \model can simulate different actions on the same scene, leading to their corresponding physical outcomes. We apply different winds (one from the left and the other from the right), and the sand castle falls in the corresponding direction.

\myparagraph{Streaming longer videos}. \model is able to create longer video streaming with a sequence of actions that go beyond the time window of baseline video models. We show results in the Appendix.

\subsection{Ablation Study}
\begin{figure}[t]
  \centering
   \includegraphics[width=\linewidth]{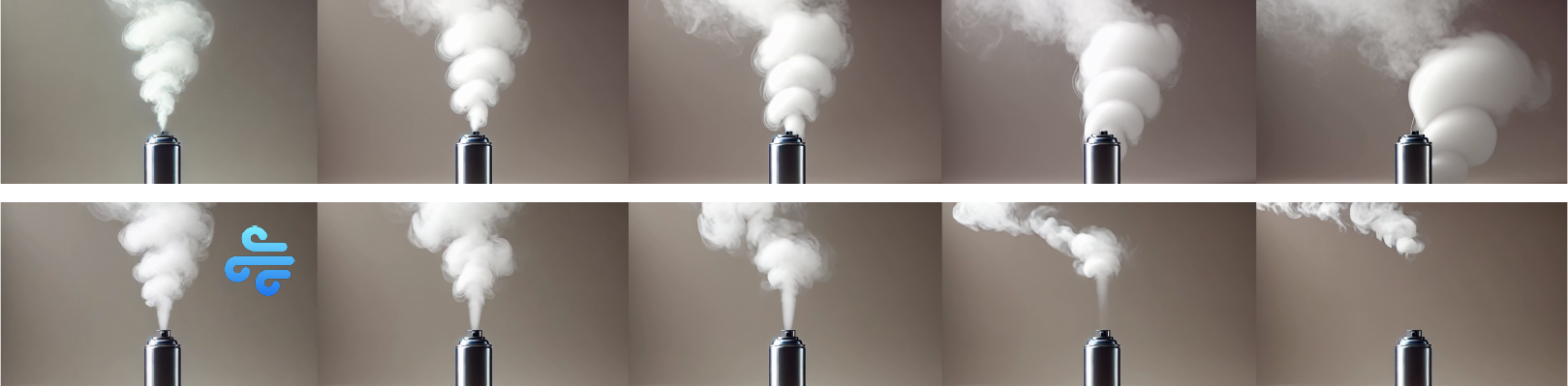}
   \caption{\textbf{Ablation} on physics simulator. \textbf{Top}: \model without the physics simulator (conditioned only on input image and text, where we specify a wind blowing from the right in the text). \textbf{Bottom}: Our full model with wind (3D force field) as action.}
   \vspace{0.2em}
   \label{fig:ablation_vid}
\end{figure}

\begin{figure}[t]
  \centering
   \includegraphics[width=\linewidth]{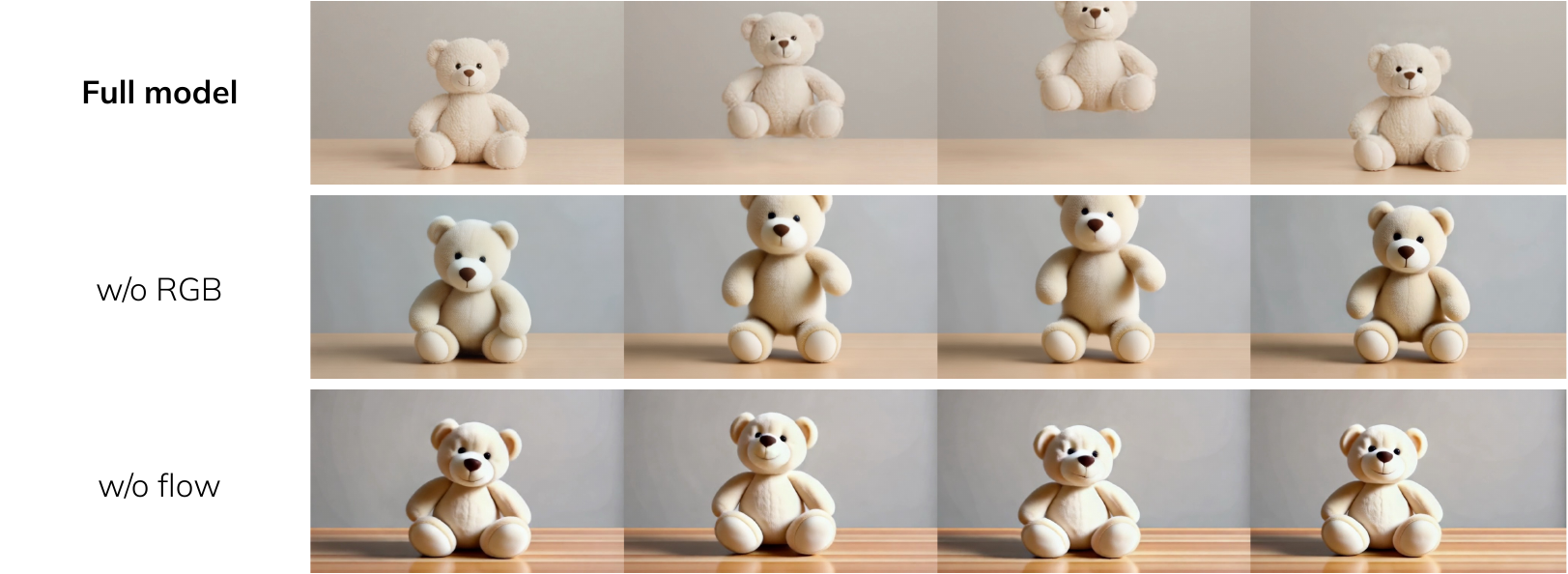}
   \caption{\textbf{Ablation} on the flow condition and RGB preview condition. We apply a 3D point force upward.}
   \label{fig:ablation}
\end{figure}

\myparagraph{Ablation on physics simulator.}
We show ablation results on the physics simulator in Figure~\ref{fig:ablation_vid}. In the top row we remove the physics simulator, so that only the prompt text provides action specification. We observe that using text alone does not lead to a plausible physical outcome (the smoke does not change direction at all) as in our full model in the bottom row.

\myparagraph{Ablation on conditioning signals.}
We show ablation results on our conditioning signals in Figure~\ref{fig:ablation}. The top row shows the synthesized video from our full model with both flow and RGB preview as conditioning signals; ``w/o RGB'' uses flow but not RGB preview; and ``w/o flow'' uses RGB but not flow. 
Without the RGB preview, the results do not adhere to the simulated overall motion. Without flow, the video model may ignore the motion signals and produce static videos. These results show that both conditioning signals are necessary.

\section{Conclusion}
\label{s:conclusion}

We present \model, a real-time physical action-conditioned video generation system. By integrating a physics simulator, we sidestep the obstacle of tokenizing continuous actions and the obstacle of curating action-video datasets. \model opens new possibilities in using video models for motion planning and AR/VR, both of which require action inputs and real-time feedback.

\myparagraph{Limitations.} Reconstructing 3D scenes can be inaccurate due to errors in depth estimation, leading to suboptimal simulation and video results. Future work may explore leveraging more reliable large reconstruction models trained on massive datasets~\cite{wang2025vggt,gslrm2024}.

\bibliographystyle{splncs04}
\bibliography{main}

\clearpage
\renewcommand\thefigure{S\arabic{figure}}
\setcounter{figure}{0}
\renewcommand\thetable{S\arabic{table}}
\setcounter{table}{0}
\renewcommand\theequation{S\arabic{equation}}
\setcounter{equation}{0}
\pagenumbering{arabic}
\renewcommand*{\thepage}{S\arabic{page}}
\setcounter{footnote}{0}
\setcounter{page}{1}
\appendix

\begin{center}
{\LARGE \textbf{Supplementary Material}}
\end{center}
\section{Additional Implementation Details}
\label{supp:impl}
\myparagraph{3D scene reconstruction.}
In our reconstruction, we leverage the Segment Anything Model 2 (SAM 2)~\cite{ravi2024sam} to segment objects from the static background, and leverage a FLUX-based inpainting model~\cite{alimama2024fluxcontrolnetinpainting} to fill the object-occluded background region. For the unprojection step, we use MoGE-2~\cite{wang2025moge2accuratemonoculargeometry} to estimate monocular depth and camera intrinsics. To reconstruct the object mesh, we utilize SAM3D~\cite{chen2025sam} as the reconstruction model, and register it to the scene coordinate space. To do this, we estimate object orientation by DUSt3R~\cite{Wang_2024_CVPR}, and then we solve for a scale $s$ and a 3D translation $\mathbf{T}$ by least square to align the two coordinate frames~\cite{li2025wonderplay}. Overall, the scene reconstruction process takes around 13.5s on a single H200 GPU.

\myparagraph{Physics simulation.}
We set all materials to be homogeneous and uniform. A rigid object is modeled as an undeformable mesh without internal links, parameterized by its mass $m$, density $\rho$, and friction coefficient $k$. We model liquid and granular substances (e.g., sand and snow) with continuum mechanics by the Material-Point-Method (MPM) solver~\cite{jiang2016material}, following prior works~\cite{xie2024physgaussian,li2025wonderplay}. They are parameterized by the density $\rho$, Young's modulus $E$, and Poisson's ratio $\nu$. For granular substance, we additionally use the friction angle $\theta$. For elastic objects, smoke, and cloth, we simulate them by the Position-Based Dynamics (PBD) solver~\cite{muller2007position,bender2015position}. PBD leverages constraints rather than physical parameters to simulate materials. Elastic objects are characterized by the stretch, bending, and volume constraints. The constraints for smoke include incompressibility~\cite{macklin2013position}. The constraints for cloth include stretch and bending compliance.

We leverage GPT-4V to estimate these materials~\cite{2023GPT4VisionSC} from the input image and they can be overridden by users. We summarize the parameters in our experiments in Table~\ref{tab:sup-params}.

\myparagraph{Video generator.}
As introduced in the main draft, our video model training is split into 3 stages.
For the first stage, our base model is VideoXFun’s~\cite{videoxfun2024} inpainting variant of Wan2.1-1.3B-T2V~\cite{wan2025}, which can refer to a starting frame and an ending frame for video inpainting. For post-training, we feed the base model the first frame and flow-wrapped noise to generate flow-conditioned videos, yielding a 1.3B I2V model that supports optical-flow conditioning.
Notably, unlike MotionStream~\cite{motionstream2025}, which uses VideoXFun’s control variant Wan2.1-1.3B-T2V~\cite{wan2025}, therefore requires additional action modules, our conditioning is directly injected into the starting noise latent, without any action embeddings or extra networks. This also simplifies the student model for distribution-matching distillation.
For the second stage of training, which is ODE regression in Self-Forcing~\cite{huang2025self}, the goal is to let the bidirectional model adapt to causal attention. We produce 2K ODE trajectories for training. The trajectories are constructed using prompts, initial frames, and flows from synthetic videos. These synthetic videos are sampled from Wan2.1-14B-T2V~\cite{wan2025}, in the hope that the model can mitigate the data bias introduced in the first stage.
For the third stage, distribution matching distillation~\cite{yin2025causvid} with Self-Forcing, we train for 600 iterations with a batch size of 64. The training data are the initial frames, prompts, and flows extracted from the same 200K synthetic videos used in the previous stage.

\myparagraph{Inference.}
We summarize our inference pipeline in Alg.~\ref{alg:pipeline}.

\begin{algorithm*}[t]
\caption{RealWonder Streaming Inference Pipeline}
\label{alg:pipeline}
\begin{algorithmic}[1]
\Require Input image $\mathbf{I}$, action stream $\{\mathbf{a}_t\}_{t=1,2,\ldots}$, text prompt $\texttt{text}$
\Ensure Video stream $\{\mathbf{V}_t\}_{t=1,2,\ldots}$
\State \textbf{// Initialization Phase}
\State $\mathcal{B}, \mathcal{O} \gets \texttt{Reconstruct3D}(\mathbf{I})$ \Comment{3D scene reconstruction}
\State $\mathbf{m} \gets \texttt{EstimateMaterials}(\mathcal{O}, \mathbf{I})$ \Comment{Material classification via VLM}
\State $\mathcal{S}_0 \gets \mathcal{B} \cup \mathcal{O}$ \Comment{Initialize scene state}
\State
\State \textbf{// Streaming Generation Loop (13.2 FPS)}
\For{$t = 1, 2, \ldots$}
    \State \textbf{// Physics Simulation}
    \State $(\mathbf{p}_{t}, \mathbf{v}_{t}) \gets \texttt{PhysicsStep}(\mathcal{S}_{t-1}, \mathbf{a}_t)$ \Comment{Simulate dynamics}
    \State $\mathcal{S}_t \gets \texttt{UpdateScene}(\mathbf{p}_t, \mathbf{v}_t)$
    \State
    \State \textbf{// Intermediate Representations}
    \State $\mathbf{F}_t \gets \Pi(\mathbf{p}_t + \Delta t \cdot \mathbf{v}_t) - \Pi(\mathbf{p}_t)$ \Comment{Compute optical flow}
    \State $\tilde{\mathbf{V}}_t \gets \texttt{RenderPointCloud}(\mathcal{S}_t)$ \Comment{Coarse RGB preview}
    \State
    \State \textbf{// Video Generation (4-step Diffusion)}
    \State $z^\mathbf{F}_t \gets \texttt{WarpNoise}(z, \mathbf{F}_t)$ \Comment{Flow-warped noise}
    \State $\mathbf{V}_{t,(3)} \gets \alpha_{(3)} \cdot \mathcal{E}(\tilde{\mathbf{V}}_t) + \sqrt{1-\alpha_{(3)}^2} \cdot z^\mathbf{F}_t$ \Comment{SDEdit mixing}
    \State $\mathbf{V}_t \gets \mathcal{G}(\texttt{text}, \mathbf{I}, \mathbf{V}_{t,(3)}, \{\mathbf{V}_j\}_{j<t})$ \Comment{Causal generation}
    \State \textbf{yield} $\mathbf{V}_t$ \Comment{Stream output frame}
\EndFor
\end{algorithmic}
\end{algorithm*}

\section{Additional Analysis}

\begin{figure}[h]
    \centering
    \includegraphics[width=\linewidth]{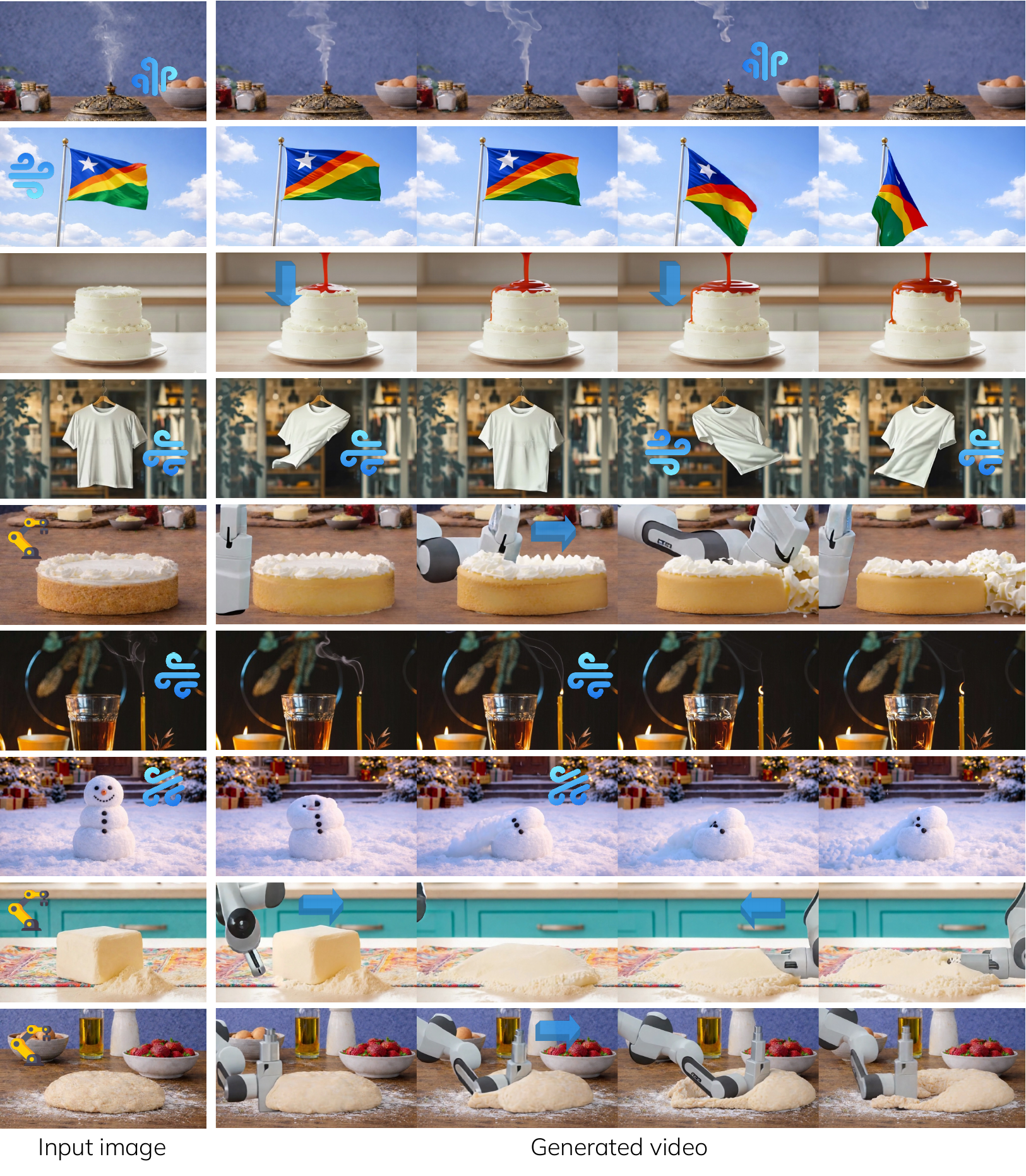}
    \caption{Additional cases across diverse materials. In the left column we show the input scene image and initial actions, where the arrow indicates 3D force, blue wind icon indicates force fields and robot icon indicates robot actions. Gravity is applied.}
    \label{fig:re_new_cases}
\end{figure}

\myparagraph{Additional Scenes.}
We include additional examples covering scenes with objects of different materials and different types of interactions, including rigid, deformable, granular, fluid, and multi-material interactions. The qualitative visualizations of these cases are shown in Figure~\ref{fig:re_new_cases}.

\begin{figure*}[h]
    \centering
    \includegraphics[width=0.8\linewidth]{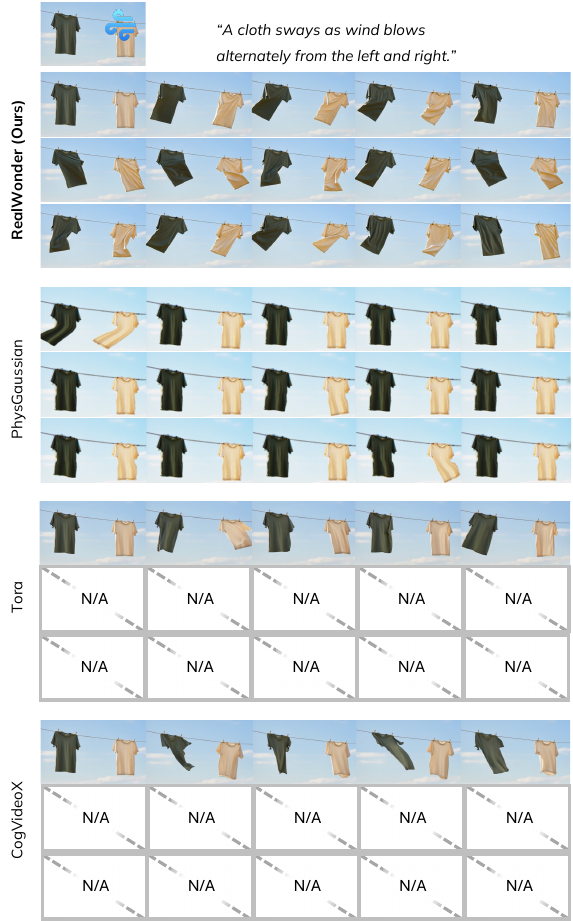}
    \caption{\textbf{Comparison to baselines on long streaming video generation}. We use "N/A" to denote frames that these methods cannot generate at later timestamps. We strongly encourage the reader to see video results in our website.}
    \label{fig:long_comp_1}
\end{figure*}

\begin{figure*}[h]
    \centering
    \includegraphics[width=0.8\linewidth]{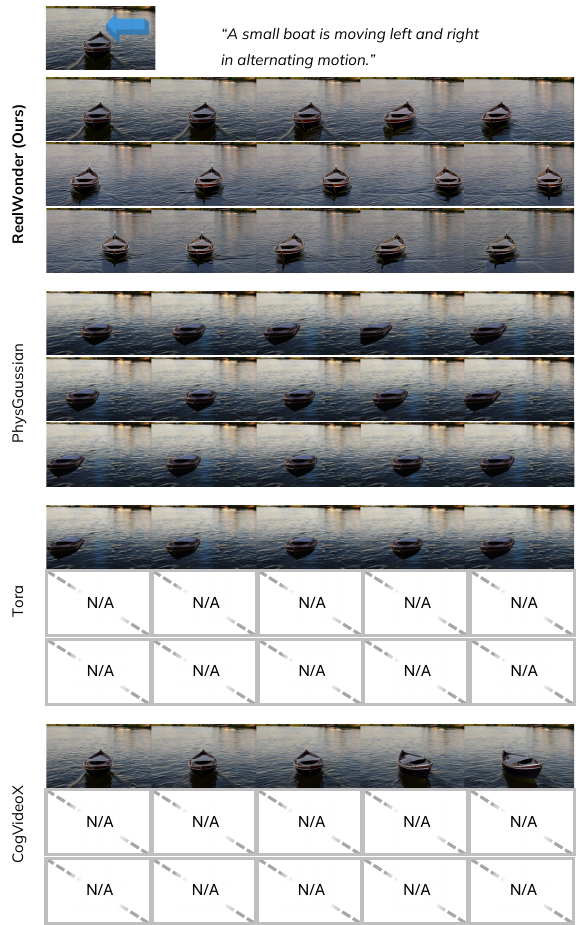}
    \caption{\textbf{Comparison to baselines on long streaming video generation}. We use "N/A" to denote frames that these methods cannot generate at later timestamps. We strongly encourage the reader to see video results in our website.}
    \label{fig:long_comp_2}
\end{figure*}

\myparagraph{Long video generation.}
Our \model naturally allows longer video streaming than a single time window. We show long video streaming examples in Figure~\ref{fig:long_comp_1} and Figure~\ref{fig:long_comp_2} compared to the baselines. While video generation-based baseline methods cannot go beyond the first time window (e.g., 5 seconds), ours allows simulating a sequence of physical actions over extended time windows. While physics-based methods like PhysGaussian~\cite{xie2024physgaussian} can also allow streaming longer videos, their visual quality quickly degenerates due to error accumulation.

\begin{figure}[t]
    \centering
    \includegraphics[width=\linewidth]{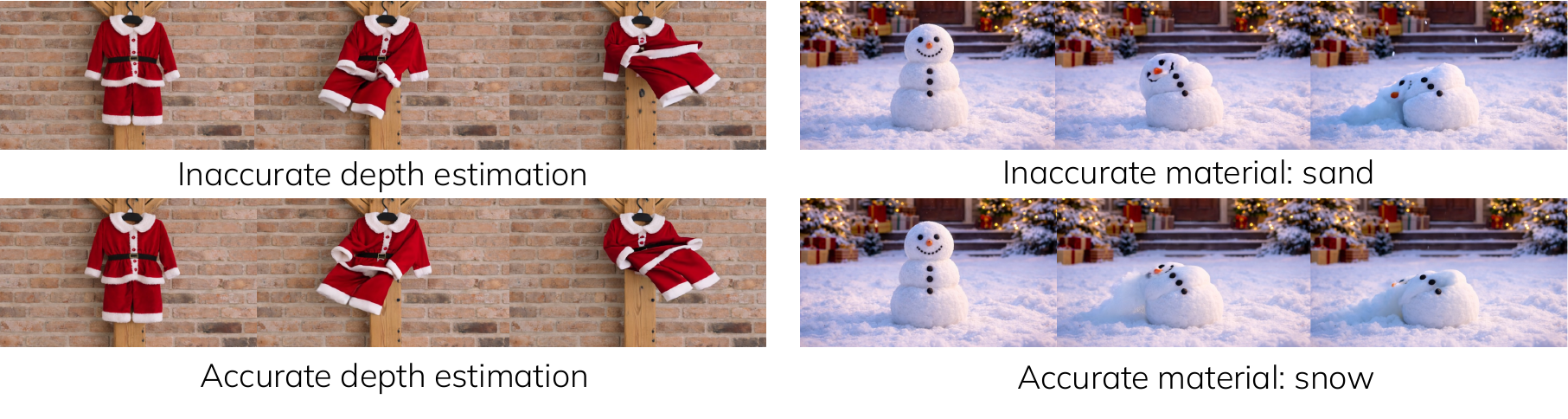}
    \caption{Consistent performance under reconstruction errors}
    \label{fig:stress_test}
\end{figure}

\myparagraph{Sensitivity to Reconstruction.}
We note that our approach does not require substantial manual effort as it is robust against reconstruction and simulation errors.
We conducted stress tests under reconstruction and material errors. Figure~\ref{fig:stress_test} shows results on cases with inaccurate depth (where we disturb object depth by $20\%$) and material (where we overwrite the estimated material from ``snow'' to ``sand''). From Figure~\ref{fig:stress_test} we observe that our method is robust against these errors in visual realism, because the video generator is robust to the minor errors in conditioning signals from the physics simulator.

\begin{figure}[t]
    \centering
    \includegraphics[width=\linewidth]{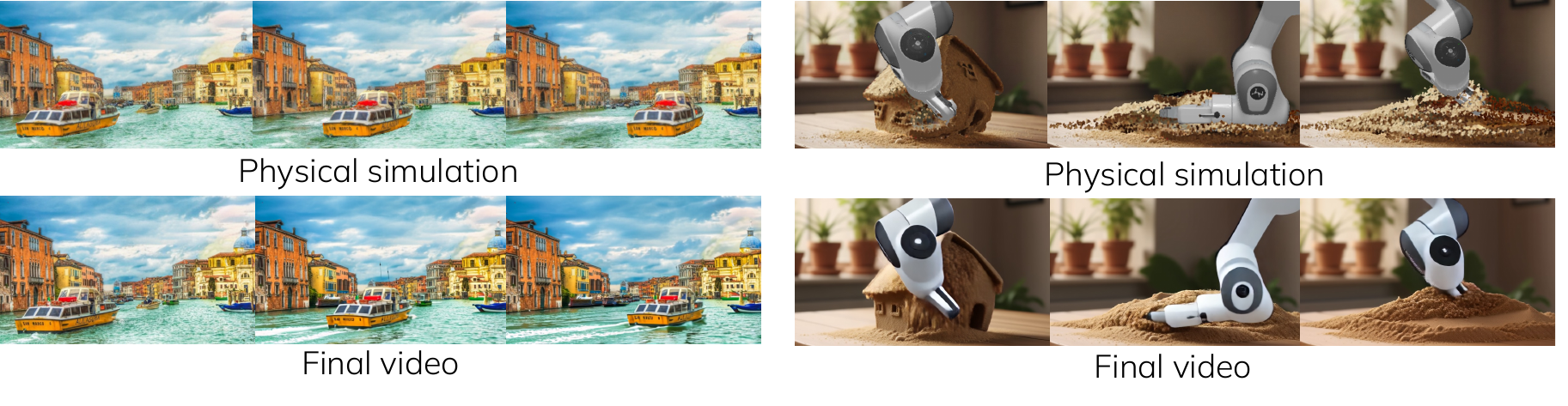}
    \caption{Left: video generator fixes ambient dynamics (i.e., the dynamics on the river). Right: video generator refines deformable object motion.}
    \label{fig:physical_plausibility}
\end{figure}

\myparagraph{Physical Plausibility.}
This work addresses the challenging problem of real-time physical action-conditioned video generation. The goal is to produce physically plausible outputs, i.e., videos that appear to be reasonable physical consequences of the input actions. Enforcing strict physical correctness, where all dynamics strictly obey physical laws, is substantially more challenging and remains an important direction for future research.
In practice, the video model can compensate for artifacts or missing dynamics in the simulator outputs. For example, Figure~\ref{fig:physical_plausibility} shows a case where the simulator provides only the motion of a boat without modeling water dynamics. The video generator synthesizes waves and ripples around the boat, producing a visually plausible result.

\myparagraph{Comparison with Teacher Video Model.}
We show a quantitative teacher-student comparison in Table~\ref{tab:supp_quan_teacher_student}. As one of our technical contributions, we show that our distilled flow-conditioned student video model successfully transfers knowledge while enabling real-time performance without quality loss.

\begin{table}[t]
\centering
\setlength{\tabcolsep}{6pt}
\caption{Quantitative comparison with teacher model}
\begin{tabular}{lccccc}
\toprule
Methods & Imaging ($\uparrow$) & Aesthetic ($\uparrow$) & Consistency ($\uparrow$) & PhysReal ($\uparrow$) \\
\midrule
Teacher & 0.713 & 0.605 & 0.271 & 0.698 \\
Student & 0.708 & 0.593 & 0.265 & 0.705 \\
\bottomrule
\end{tabular}
\label{tab:supp_quan_teacher_student}
\end{table}

\begin{table}[t]
\begin{center}
\caption{Simulation parameters and default values}\label{tab:sup-params}
\begin{tabular}{lc}
\toprule
 Parameter & Default Value \\ \midrule
\textbf{General simulation} &  \\
 Step time & $1e^{-2}$ \\
 Sub-steps number & $10$ \\
 Sampled particle size & $1e ^{-2}$ \\
 Gravity & $(0, 0, -9.8)$ \\

\midrule
 \textbf{Rigid body solver} &  \\
 friction coefficient & $0.1$ \\

\midrule
\textbf{MPM solver} & \\
Grid density & $64$ \\

Liquid material Young's modulus & $1e^7$ \\
Liquid material Poisson's ratio & $0.2$ \\

Granular material Young's modulus & $1e^6$ \\
Granular material Poisson's ratio & $0.2$ \\
Granular material Friction angle & $45$ \\
\midrule

\textbf{PBD solver} & \\
Elastic stretch compliance: &0 \\
Elastic bending compliance: &0 \\
Elastic volume compliance: &0 \\
Elastic stretch relaxation: &0.3 \\
Elastic bending relaxation: &0.3 \\
Elastic volume relaxation: &0.1 \\
Cloth material stretch compliance & $1e^{-7}$ \\
Cloth material bending compliance & $1e^{-5}$ \\
Smoke material viscosity coefficient & $0.1$ \\

\bottomrule
\end{tabular}
\end{center}
\end{table}

\end{document}